\documentclass[letterpaper]{article} 
\usepackage{amsmath}

\usepackage{aaai23}  
\usepackage{algorithm}
\usepackage{algorithmicx}
\usepackage[noend]{algpseudocode}
\usepackage{times}  
\usepackage{helvet}  
\usepackage{courier}  
\usepackage[hyphens]{url}  
\usepackage{graphicx} 
\usepackage{natbib}  
\usepackage{caption} 
\usepackage{amssymb}
\usepackage{mathtools}
\usepackage{amsthm}
\usepackage{xcolor} 
\usepackage{booktabs}
\usepackage{multirow}
\usepackage{pifont}
\newcommand{\cmark}{\ding{51}}%
\newcommand{\xmark}{\ding{55}}%
\newcommand*{\balancecolsandclearpage}{%
  \close@column@grid
  \cleardoublepage
  \twocolumngrid
}
\usepackage[skip=2pt]{caption}  
\usepackage{newfloat}
\usepackage{listings}
\DeclareCaptionStyle{ruled}{labelfont=normalfont,labelsep=colon,strut=off} 
\floatstyle{ruled}
\newfloat{listing}{tb}{lst}{}
\floatname{listing}{Listing}
\title{Sparse Coding in a Dual Memory System for Lifelong Learning}
\author{
    Fahad Sarfraz\equalcontrib\textsuperscript{\rm 1},
    Elahe Arani\equalcontrib\textsuperscript{\rm 1,2},
    Bahram Zonooz\textsuperscript{\rm 1,2} \\
}
\affiliations{
    \textsuperscript{\rm 1}Advanced Research Lab, NavInfo Europe, The Netherlands\\
    \textsuperscript{\rm 2}Department of Mathematics and Computer Science, Eindhoven University of Technology, The Netherlands\\
\footnotesize\texttt{fahad.sarfraz@navinfo.eu, elahe.arani@tue.nl, bahram.zonooz@gmail.com}
}

\usepackage{bibentry}

\begin{document}

\maketitle

\begin{abstract}

Efficient continual learning in humans is enabled by a rich set of neurophysiological mechanisms and interactions between multiple memory systems. The brain efficiently encodes information in non-overlapping sparse codes, which facilitates the learning of new associations faster with controlled interference with previous associations. To mimic sparse coding in DNNs, we enforce activation sparsity along with a dropout mechanism which encourages the model to activate similar units for semantically similar inputs and have less overlap with activation patterns of semantically dissimilar inputs. This provides us with an efficient mechanism for balancing the reusability and interference of features, depending on the similarity of classes across tasks. Furthermore, we employ sparse coding in a multiple-memory replay mechanism. Our method maintains an additional long-term semantic memory that aggregates and consolidates information encoded in the synaptic weights of the working model. 
Our extensive evaluation and characteristics analysis show that equipped with these biologically inspired mechanisms, the model can further mitigate forgetting\footnote{Code available at \textcolor{blue}{\url{https://github.com/NeurAI-Lab/SCoMMER}}}.


\end{abstract}

\section{Introduction}

The ability to continually acquire, consolidate, and retain knowledge is a hallmark of intelligence. Particularly, as we look to deploy deep neural networks (DNNs) in the real world, it is essential that learning agents continuously interact and adapt to the ever-changing environment. However, standard DNNs are not designed for lifelong learning and exhibit catastrophic forgetting of previously learned knowledge~\citep{mccloskey1989catastrophic} when required to learn tasks sequentially from a stream of data~\citep{mccloskey1989catastrophic}.


The core challenge in continual learning (CL) in DNNs is to maintain an optimal balance between plasticity and the stability of the model. Ideally, the model should be stable enough to retain previous knowledge while also plastic enough to acquire and consolidate new knowledge.
Catastrophic forgetting in DNNs can be attributed to the lack of stability, and multiple approaches have been proposed to address it.
Among them, \textit{Rehearsal-based} methods,~\citep{riemer2018learning,aljundi2019gradient} which aim to reduce forgetting by continual rehearsal of previously seen tasks, have proven to be an effective approach in challenging CL tasks \citep{farquhar2018towards}. They attempt to approximate the joint distribution of all the observed tasks by saving samples from previous tasks in a memory buffer and intertwine the training of the new task with samples from memory. However, due to the limited buffer size, it is difficult to approximate the joint distribution with the samples alone. There is an inherent imbalance between the samples of previous tasks and the current task. This results in the network update being biased towards the current task, leading to forgetting and recency bias in predictions. Therefore, more information from the previous state of the model is needed to better approximate the joint distribution and constrain the update of the model to preserve the learned knowledge. However, it is still an open question what the optimal information is for replay and how to extract and preserve it.

The human brain provides an existence proof for successful CL in complex dynamic environments without intransigence or forgetting. Therefore, it can provide insight into the design principles and mechanisms that can enable CL in DNNs. The human brain maintains a delicate balance between stability and plasticity through a complex set of neurophysiological mechanisms \citep{parisi2019continual,zenke2017continual} and the effective use of multiple memory systems \citep{hassabis2017neuroscience}. In particular, evidence suggests that the brain employs \textit{Sparse Coding,} that the neural code is characterized by strong activations of a relatively small set of neurons. The efficient utilization of sparsity for information representation enables new associations to be learned faster with controlled interference with previous associations while maintaining sufficient representation capacity. Furthermore, complementary learning systems (CLS) theory posits that effective learning requires two complementary learning systems. The hippocampus rapidly encodes episodic information into non-overlapping representations, which are then gradually consolidated into the structural knowledge representation in the neocortex through the replay of neural activities.


\begin{figure*}
    \centering
    \includegraphics[width=.9\linewidth]{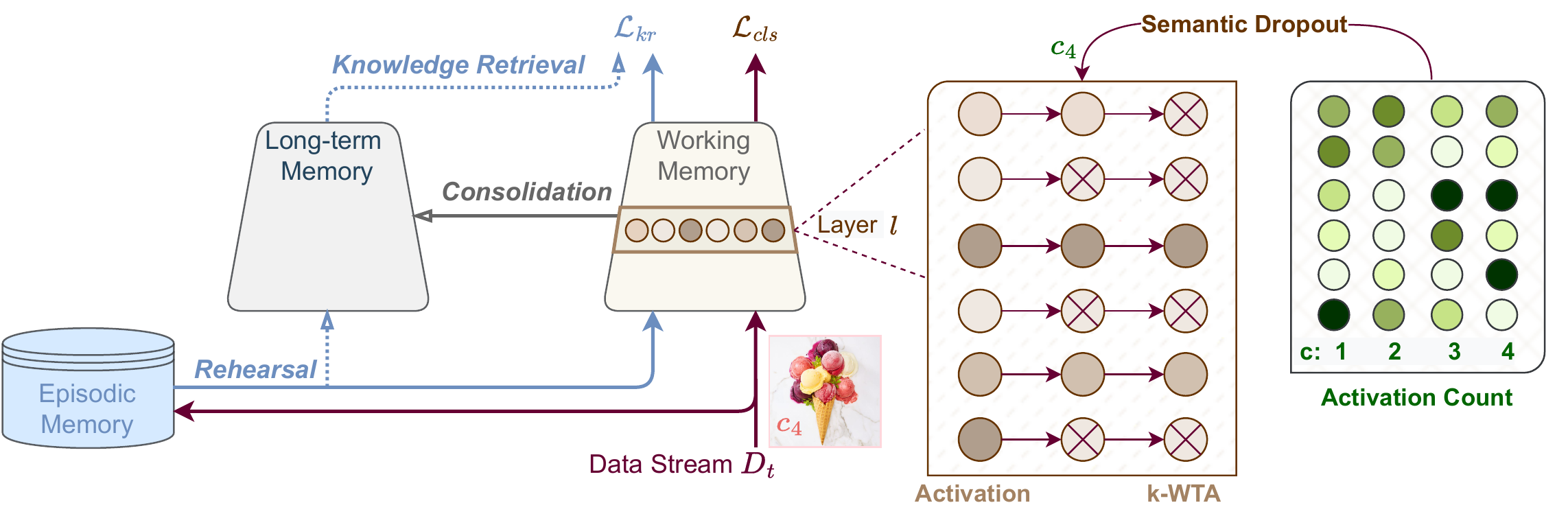}
    \caption{SCoMMER employs sparse coding in a multi-memory experience replay mechanism. In addition to the instance-based episodic memory, we maintain a long-term memory that consolidates the learned knowledge in the working memory throughout training. The long-term memory interacts with the episodic memory to enforce consistency in the functional space of working memory through the knowledge retrieval loss. To mimic sparse coding in the brain, we enforce activation sparsity along with semantic dropout, whereby the model tracks the class-wise activations during training and utilizes them to enforce sparse code, which encourages the model to activate similar units for semantically similar inputs. Schematic shows how the activations from layer $l$ are propagated to the next layer. Darker shades indicate higher values. Given a sample from class 4, semantic dropout retains the units with higher activation counts for the class, and top-k remaining (here 2) units with higher activations are propagated to the next layer. This enables the network to form semantically conditioned subnetworks and mitigate forgetting.}
    \label{fig:scommer}
\end{figure*}

Inspired by these mechanisms in the brain, we hypothesize that employing a mechanism to encourage sparse coding in DNNs and mimic the interplay of multiple memory systems can be effective in maintaining a balance between stability and plasticity. To this end, we propose a multi-memory experience replay mechanism that employs sparse coding, SCoMMER. We enforce activation sparsity along with a complementary dropout mechanism, which encourages the model to activate similar units for semantically similar inputs while reducing the overlap with activation patterns of semantically dissimilar inputs. The proposed semantic dropout provides us with an efficient mechanism to balance the reusability and interference of features depending on the similarity of classes across tasks. Furthermore, we maintain additional long-term semantic memory that aggregates the information encoded in the synaptic weights of the working memory. Long-term memory interacts with episodic memory to retrieve structural knowledge from previous tasks and facilitates information consolidation by enforcing consistency in functional space. 

Our empirical evaluation on challenging CL settings and characteristic analysis show that equipping the model with these biologically inspired mechanisms can further mitigate forgetting and effectively consolidate information across the tasks. Furthermore, sparse activations in conjunction with semantic dropout in SCoMMER leads to the emergence of subnetworks, enables efficient utilization of semantic memory, and reduces the bias towards recent tasks.

\section{Related Work}
The different approaches to address the problem of catastrophic forgetting in CL can be broadly divided into three categories: \textit{Regularization-based} methods regularize the update of the model in the parameter space~\citep{farajtabar2020orthogonal, kirkpatrick2017overcoming, ritter2018online, zenke2017continual} or the functional space~\citep{rannen2017encoder,li2017learning}, \textit{Dynamic architecture} expands the network to dedicate a distinct set of parameters to each task, and \textit{Rehearsal-based} methods~\citep{riemer2018learning,aljundi2019gradient} mitigate forgetting by maintaining an episodic memory buffer and continual rehearsal of samples from previous tasks. Among these, our method focuses on rehearsal-based methods, as it has proven to be an effective approach in challenging continual learning scenarios~\citep{farquhar2018towards}. The base method, Experience Replay (ER)~\citep{riemer2018learning} interleaves the training of the current task with the memory sample to train the model on the approximate joint distribution of tasks. Several studies focus on the different aspects of rehearsal: memory sample selection~\cite{lopez2017gradient,isele2018selective}, sample retrieval from memory~\cite{aljundi2019online} and what information to extract and replay from the previous model~\cite{li2017learning,ebrahimi2020remembering,bhat2022consistency}.

Dark Experience Replay (DER++) samples the output logits along with the samples in the memory buffer throughout the training trajectory and applies a consistency loss on the update of the model. Recently, CLS theory has inspired a number of approaches that utilize multiple memory systems~\cite{wang2022dualprompt,wang2022learning,pham2021dualnet} and show the benefits of multiple systems in CL. CLS-ER~\cite{arani2021learning} mimics the interplay between fast and slow learning systems by maintaining two additional semantic memories that aggregate the weights of the working model at different timescales using an exponential moving average. Our method enforces sparse coding for efficient representation and utilization of multiple memories. 

\section{Methodology}
We first provide an overview of motivation from biological systems before formally introducing the different components of the proposed approach.

\subsection{Continual Learning in the Biological System}
Effective CL in the brain is facilitated by a complex set of mechanisms and multiple memory systems. Information in the brain is represented by neural activation patterns, which form a neural code~\cite{foldiak2008sparse}. Specifically, evidence suggests that the brain employs \textit{Sparse Coding,} in which sensory events are represented by strong activations of a relatively small set of neurons. A different subset of neurons is used for each stimulus~\cite{foldiak2003sparse,barth2012experimental}. There is a correlation between these sparse codes~\cite{lehky2021pseudosparse} that could capture the similarity between different stimuli. 
Sparse codes provide several advantages: they enable faster learning of new associations with controlled interference with previous associations and allow efficient maintenance of associative memory while retaining sufficient representational capacity.

Another salient feature of the brain is the strong differentiation and specialization of the nervous systems~\cite{hadsell2020embracing}. There is evidence for modularity in biological systems, which supports functional specialization of brain regions~\cite{kelkar2018evidence} and reduces interference between different tasks. Furthermore, the brain is believed to utilize multiple memory systems~\cite{atkinson1968human,mcclelland1995there}. Complementary learning systems (CLS) theory states that efficient learning requires at least two complementary systems. The instance-based hippocampal system rapidly encodes new episodic events into non-overlapping representations, which are then gradually consolidated into the structured knowledge representation in the parametric neocortical system. Consolidation of information is accompanied by replay of the neural activities that accompanied the learning event. 

The encoding of information into efficient sparse codes, the modular and dynamic processing of information, and the interplay of multiple memory systems might play a crucial role in enabling effective CL in the brain. Therefore, our method aims to incorporate these components in ANNs.

\subsection{Sparse coding in DNNs}
The sparse neural codes in the brain are in stark contrast to the highly dense connections and overlapping representations in standard DNNs which are prone to interference. In particular, for CL, sparse representations can reduce the interference between different tasks and therefore result in less forgetting, as there will be fewer task-sensitive parameters or fewer effective changes to the parameters~\cite{abbasi2022sparsity,iyer2021avoiding}. Activation sparsity can also lead to the natural emergence of modules without explicitly imposing architectural constraints~\cite{hadsell2020embracing}.
Therefore, to mimic sparse coding in DNNs, we enforce activation sparsity along with a complementary semantic dropout mechanism which encourages the model to activate similar units for semantically similar samples.

\subsubsection{Sparse Activations:}
To enforce the sparsity in activations, we employ the k-winner-take-all (k-WTA) activation function~\cite{maass2000computational}. k-WTA only retains the top-k largest values of an $N \times 1$ input vector and sets all the others to zero before propagating the vector to the next layer of the network. 
Importantly, we deviate from the common implementation of k-WTA in convolutional neural networks (CNNs) whereby the activation map of a layer ($C \times H \times W$ tensor where $C$ is the number of channels and $H$ and $W$ are the spatial dimensions) is flattened into a long $CHW \times 1$ vector input and the k-WTA activation is applied similar to the fully connected network~\cite{xiao2019enhancing,ahmad2019can}. We believe that this implementation does not take into account the functional integrity of an individual convolution filter as an independent feature extractor and does not lend itself to the formation of task-specific subnetworks with specialized feature extractors. Instead, we assign an activation score to each filter in the layer by taking the absolute sum of the corresponding activation map and select the top-k filters to propagate to the next layer.

Given the activation map, we flatten the last two dimensions and assign a score to each filter by taking the absolute sum of the activations. Based on the sparsity ratio for each layer, the activation maps of the filters with higher scores are propagated to the next layers, and the others are set to zero. This enforces global sparsity, whereby each stimulus is processed by only a selected set of convolution filters in each layer, which can be considered as a subnetwork. We also consider each layer's role when setting the sparsity ratio. The earlier layers have a lower sparsity ratio as they learn general features, which can enable higher reusability, and forward transfer to subsequent tasks use a higher sparsity for later layers to reduce the interference between task-specific features.

\begin{algorithm*}[t]
\caption{SCoMMER Algorithm for Sparse Coding in Multiple-Memory Experience Replay System}
\label{algo:SCoMMER}
\begin{algorithmic}[1]
\Statex {\bf Input:} {data stream $\mathcal{D}$; learning rate $\eta$; consistency weight $\gamma$; update rate $r$ and decay parameter $\alpha$, dropout rates $\pi_{h}$ and $\pi_{s}$}

\Statex {\bf Initialize: }{$\theta_s = \theta_w$}
\Statex {}{$\mathcal{M}\xleftarrow{}\{\}$} 

\For{$\mathcal{D}_t \in \mathcal{D}$}
\While {Training}
\State Sample training data: {$(x_t,y_t) \sim \mathcal{D}_t$} and {$(x_m,y_m) \sim \mathcal{M}$, and interleave $x \leftarrow (x_t,x_m)$} 

\State {Retrieve structural knowledge: $\mathcal{Z}_{s} \leftarrow f(x_m; \theta_s) $} 
\State {Evaluate overall loss loss: $\mathcal{L} = \mathcal{L}_{ce}(f(x; \theta_w), y) + \gamma \mathcal{L}_{kr}(f(x_m; \theta_w), \mathcal{Z}_{s})$  ~~(Eq. \ref{eq:loss})} 

\State {Update working memory: $\theta_w \xleftarrow{} \theta_w - \eta \nabla_{\theta_w}\mathcal{L}$}

\State {Aggregate knowledge: $\theta_s \leftarrow \alpha \theta_s+(1-\alpha)~\theta_w,~~~if~~ r > a \sim U(0,1)$ ~~(Eq. \ref{eq:sem_update})}

\State {Update episodic memory: $\mathcal{M} \xleftarrow{} \text{Reservoir} (\mathcal{M}, (x_t,y_t))$}

\State {After $\mathcal{E}_h$ epochs, update semantic dropout probabilities at the end of each epoch: $P_{s}$ ~~ (Eq.~\ref{eq:semantic_dropout})}

\EndWhile
\State {Update heterogeneous dropout probabilities: $P_h$ ~~ (Eq.~\ref{eq:het_dropout})}
\EndFor
\Statex \Return{$\theta_s$}
\end{algorithmic}
\end{algorithm*}

\subsubsection{Semantic Dropout:} While the k-WTA activation function enforces the sparsity of activation for each stimulus, it does not encourage semantically similar inputs to have similar activation patterns and reduce overlap with semantically dissimilar inputs. To this end, we employ a complementary \textit{Semantic Dropout} mechanism, which controls the degree of overlap between neural activations between samples belonging to different tasks while also encouraging the samples belonging to the same class to utilize a similar set of units. We utilize two sets of activation trackers: \textit{global activity counter}, $\mathcal{A}_g \in \mathbb{R}^N$, counts the number of times each unit has been activated throughout training, whereas \textit{class-wise activity counter}, $\mathcal{A}_s \in \mathbb{R}^{C \times N}$, tracks the number of times each unit has been active for samples belonging to a particular class. $N$ and $C$ denote the total number of units and classes, respectively. For each subsequent task, we first employ Heterogeneous Dropout~\cite{abbasi2022sparsity} to encourage the model to learn the new classes by using neurons that have been less active for previously seen classes by setting the probability of a neuron being dropped to be inversely proportional to its activation counts. Concretely, let $[\mathcal{A}_{g}^l]_j$ denote the number of times that the unit $j$ in layer $l$ has been activated after learning $t$ sequential tasks. For learning the new classes in task $t$+1, the probability of retaining this unit is given by:
\begin{equation} \label{eq:het_dropout}
    [P_h^l]_j = exp(\frac{-[\mathcal{A}_{g}^l]_j}{\max_{i}{[\mathcal{A}_{g}^l]_i}}\pi_{h})
\end{equation}
where $\pi_{h}$ controls the strength of dropout with larger values leading to less overlap between representations. We then allow the network to learn with the new task with heterogeneous dropout in place of a fixed number of epochs, $\mathcal{E}_{h}$. During this period, we let the class-wise activations emerge and then employ \textit{Semantic Dropout}. It encourages the model to utilize the same set of units by setting the probability of retention of a unit for each class $c$ as proportional to the number of times it has been activated for that class so far:
\begin{equation} \label{eq:semantic_dropout}
    [P_s^l]_{c,j} = 1 - exp(\frac{-[\mathcal{A}_s^l]_{c,j}}{\max_{i}{[\mathcal{A}_s^l]_{c,i}}} \pi_{s})
\end{equation}
where $\pi_{s}$ controls the strength of dropout. The probabilities for semantic dropout are updated at the end of each epoch to enforce the emerging pattern. This provides us with an efficient mechanism for controlling the degree of overlap in representations as well as enabling context-specific processing of information which facilitates the formation of semantically conditioned subnetworks. Activation sparsity, together with semantic dropout, also provides us with an efficient mechanism for balancing the reusability and interference of features depending on the similarity of classes across the tasks.

\subsection{Multiple Memory Systems}
Inspired by the interaction of multiple memory systems in the brain, in addition to a fixed-size instance-based episodic memory, our method builds a long-term memory that aggregates the learned information in the working memory.

\subsubsection{Episodic Memory:}
Information consolidation in the brain is facilitated by replaying the neural activation patterns that accompanied the learning event. To mimic this mechanism, we employ a fixed-size episodic memory buffer, which can be thought of as a very primitive hippocampus. The memory buffer is maintained with \textit{Reservoir Sampling,} \citep{vitter1985random} which aims to match the distribution of the data stream by assigning an equal probability to each incoming sample.

\subsubsection{Long-Term Memory:}
We aim to build a long-term semantic memory that can consolidate and accumulate the structural knowledge learned in the working memory throughout the training trajectory. The knowledge acquired in DNNs resides in the learned synaptic weights~\citep{krishnan2019biologically}. Hence, progressively aggregating the weights of the working memory ($\theta_w$) as it sequentially learns tasks allows us to consolidate the information efficiently.
To this end, we build long-term memory ($\theta_s$) by taking the exponential moving average of the working memory weights in a stochastic manner (which is more biologically plausible~\cite{arani2021noise}), similar to \citep{arani2021learning}:
\begin{equation} \label{eq:sem_update}
    \theta_s \leftarrow \alpha \theta_s+(1-\alpha)~\theta_w,~~~~~if~~ r>a \sim U(0,1)
\end{equation}
where $\alpha$ is the decay parameter and $r$ is the update rate.

\begin{table*}[tb]
\caption{Comparison on different CL settings. The baseline results for S-CIFAR100 and GCIL are from \cite{arani2021learning}.}
\label{tab:all-datasets}
\centering
\begin{tabular}{@{\extracolsep{4pt}}llcccccc@{}}
\toprule
\multirow{2}{*}{{Buffer}} & \multirow{2}{*}{{Method}} & \multicolumn{2}{c}{\textbf{S-CIFAR10}} & \multicolumn{2}{c}{\textbf{S-CIFAR100}} & \multicolumn{2}{c}{\textbf{GCIL}} \\ \cmidrule{3-4} \cmidrule{5-6} \cmidrule{7-8}
 &  &  {Class-IL} & {Task-IL} &  {Class-IL} & {Task-IL} & {Unif} & {Longtail} \\ \midrule
 
\multirow{2}{*}{–} & JOINT & 92.20\tiny±0.15 & 98.31\tiny±0.12 & 70.62\tiny±0.64 & 86.19\tiny±0.43  & 58.36\tiny±1.02 & 56.94\tiny±1.56 \\
 & SGD & 19.62\tiny±0.05 & 61.02\tiny±3.33 & 17.58\tiny±0.04 & 40.46\tiny±0.99  & 12.67\tiny±0.24 & 22.88\tiny±0.53 \\  \midrule
 
\multirow{4}{*}{200} & ER & 44.79\tiny±1.86 & 91.19\tiny±0.94 & 21.40\tiny±0.22 & 61.36\tiny±0.39 & 16.40\tiny±0.37 & 19.27\tiny±0.77 \\ 
 & DER++ & 64.88\tiny±1.17 & 91.92\tiny±0.60 & 29.60\tiny±1.14 & 62.49\tiny±0.78 & 18.84\tiny±0.60 & 26.94\tiny±1.27 \\ 
 & CLS-ER & 66.19\tiny±0.75 & \textbf{93.90}\tiny±0.60 & 35.23\tiny±0.86 & 67.34\tiny±0.79 & 25.06\tiny±0.81 & 28.54\tiny±0.87 \\
 & SCoMMER & \textbf{69.19}\tiny±0.61 & 93.20\tiny±0.10 & \textbf{40.25}\tiny±0.05 & \textbf{69.39}\tiny±0.43 & \textbf{30.84}\tiny±0.80 & \textbf{29.08}\tiny±0.31 \\ 
 \midrule

\multirow{4}{*}{500} & ER & 57.74\tiny±0.27 & 93.61\tiny±0.27 & 28.02\tiny±0.31  & 68.23\tiny±0.16 & 28.21\tiny±0.69 & 20.30\tiny±0.63 \\ 
 & DER++ & 72.70\tiny±1.36 & 93.88\tiny±0.50 & 41.40\tiny±0.96  & 70.61\tiny±0.11 & 32.92\tiny±0.74 & 25.82\tiny±0.83 \\ 
 & CLS-ER & \textbf{75.22}\tiny±0.71 & \textbf{94.94}\tiny±0.53 & 47.63\tiny±0.61 & 73.78\tiny±0.86 & 36.34\tiny±0.59 & 28.63\tiny±0.68 \\
 & SCoMMER & 74.97\tiny±1.05 & 94.36\tiny±0.06 & \textbf{49.63}\tiny±1.43  & \textbf{75.49}\tiny±0.43 & \textbf{36.87}\tiny±0.36 & \textbf{35.20}\tiny±0.21 \\
\bottomrule
\end{tabular}
\end{table*}

Long-term memory builds structural representations for generalization and mimics the slow acquisition of structured knowledge in the neocortex, which can generalize well across tasks. The long-term memory then interacts with the instance-level episodic memory to retrieve structural relational knowledge~\cite{sarfraz2021knowledge} for the previous tasks encoded in the output logits. Consolidated logits are then utilized to enforce consistency in the functional space of the working model. This facilitates the consolidation of information by encouraging the acquisition of new knowledge while maintaining the functional relation of previous knowledge and aligning the decision boundary of working memory with long-term memory. 

\subsection{Overall Formulation}
Given a continuous data stream $\mathcal{D}$ containing a sequence of tasks ($\mathcal{D}_1, \mathcal{D}_2, .., \mathcal{D}_T$), the CL task is to learn the joint distribution of all the observed tasks without the availability of task labels at test time. Our proposed method, SCoMMER, involves training a working memory $\theta_w$, and maintains an additional long-term memory $\theta_s$ and an episodic memory $\mathcal{M}$. The long-term memory is initialized with the same parameters as the working memory and has the same sparsity constraints. Therefore, long-term memory aggregates the weights of working memory. We initialize heterogeneous dropout probabilities $\pi_h$ randomly to set the probability of retention of a fraction of units to 1 and others to 0 so that the first task is learned using a few, but sufficient units and the remaining can be utilized to learn subsequent tasks.

\begin{figure}[t]
    \centering
    \includegraphics[width=1\linewidth]{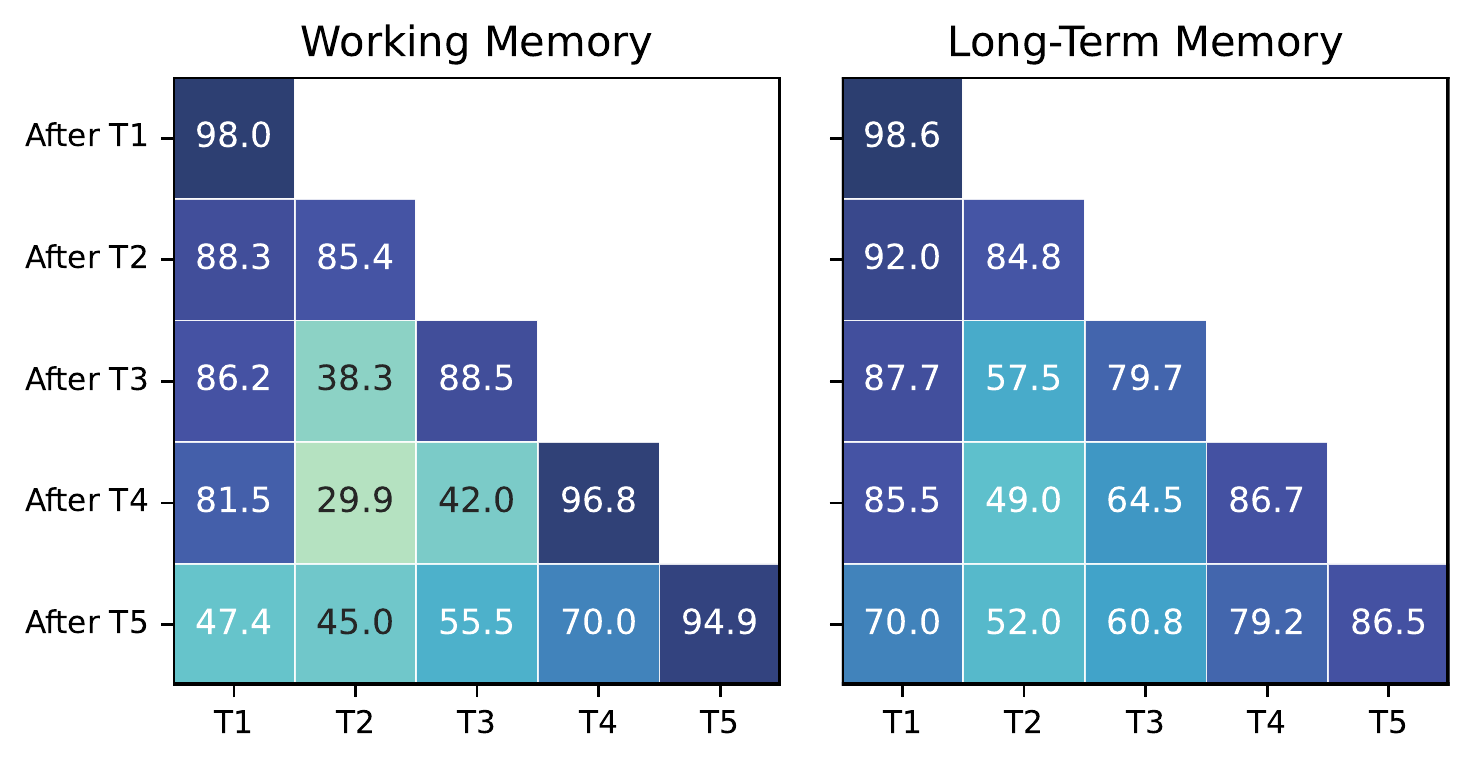}
    \caption{Task-wise performance of working memory and the long-term memory. The long-term memory effectively aggregates knowledge encoded in the working memory and generalizes well across the tasks.}
    \label{fig:comp_perf}
\end{figure}

During each training step, we interleave the batch of samples from the current task $x_t \sim \mathcal{D}_t$, with a random batch of exemplars from episodic memory $x_m \sim \mathcal{M}$. Working memory is trained with a combination of cross-entropy loss on the interleaved batch $x\leftarrow(x_t,x_b)$, and knowledge retrieval loss on the exemplars.
Thus, the overall loss is given by:
\begin{equation} \label{eq:loss}
    \mathcal{L} =\mathcal{L}_{ce}(f(x; \theta_w), y) + \gamma \mathcal{L}_{kr}(f(x_m; \theta_w), f(x_m; \theta_s))
\end{equation}
where $\gamma$ controls the strength of the enforcement of consistency, and mean-squared error loss is used for $\mathcal{L}_{kr}$. The training step is followed by stochastically updating the long-term memory (Eq.~\ref{eq:sem_update}). The semantic dropout and heterogeneous dropout probabilities are updated at the end of each epoch and task, respectively (using Eqs. \ref{eq:het_dropout} and \ref{eq:sem_update}).
We use long-term memory for inference, as it aggregates knowledge and generalizes well across tasks (cf. Figure~\ref{fig:comp_perf}). Agorithm~\ref{algo:SCoMMER} provides further training details.


\section{Evaluation Protocol}
To gauge the effectiveness of SCoMMER in tackling the different challenges faced by a lifelong learning agent, we consider multiple CL settings that test different aspects of the model.

\textbf{Class-IL} presents a challenging CL scenario where each task presents a new set of disjoint classes, and the model must learn to distinguish between all the classes seen so far without the availability of task labels at the test time. It requires the model to effectively consolidate information across tasks and learn generalizable features that can be reused to acquire new knowledge. \textbf{Generalized Class-IL (GCIL)}~\cite{mi2020generalized} extends the Class-IL setting to more realistic scenarios where the agent has to learn an object over multiple recurrences spread across tasks and tackle the challenges of class imbalance and a varying number of classes in each task. GCIL utilizes probabilistic modeling to sample the number of classes, the appearing classes, and their sample sizes. Details of the datasets used in each setting are provided in the Appendix. Though our method does not utilize separate classification heads or subnets, for completion, we also evaluate the performance under the Task-IL setting, where the model has access to the task labels at inference. In this setting, we use the task label to select the subset of output logits to select from.

\begin{figure*}[t]
    \centering
    \includegraphics[width=.95\linewidth]{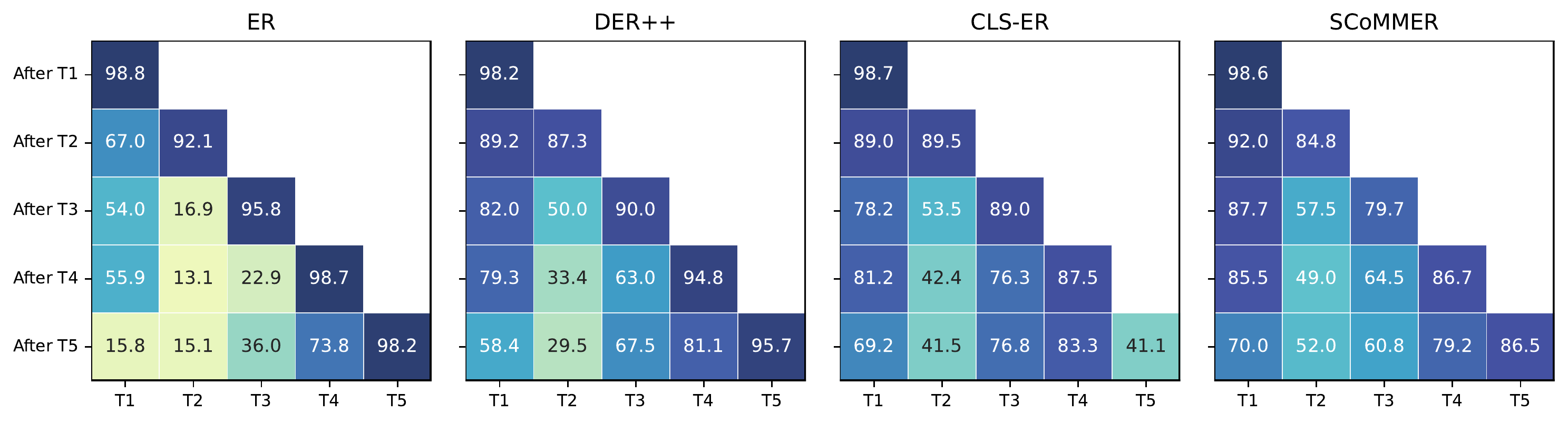}
    \caption{Task-wise performance of different methods. The heatmaps provide the test set of each task (x-axis) evaluated at the end of each sequential learning task (y-axis). SCoMMER retains the performance of earlier tasks better without compromising on the current task.}
    \label{fig:task_perf}
\end{figure*}

\section{Empirical Evaluation}
We compare SCoMMER with state-of-the-art rehearsal-based methods across different CL settings under uniform experimental settings (details provided in Appendix). \textit{SGD} provides the lower bound with standard training on sequential tasks, and \textit{JOINT} gives the upper bound on performance when the model is trained on the joint distribution. 

Table~\ref{tab:all-datasets} shows that SCoMMER provides performance gains in the majority of the cases and demonstrates the effectiveness of our approach under varying challenging CL settings. In particular, it provides considerable improvement under low buffer size settings, which suggests that our method is able to mitigate forgetting with fewer samples from previous tasks. The performance gains over CLS-ER, which employs two semantic memories, show that sparse coding in our method enables the effective utilization of a single semantic memory. In particular, the gains in the GCIL setting, where the agent has to face the challenges of class imbalance and learn over multiple occurrences of objects, alludes to several advantages of our method. Our proposed semantic dropout in conjunction with sparse activations enables the model to reuse the sparse code associated with the recurring object and learn better representations with the additional samples by adapting the corresponding subset of filters. Furthermore, compared to the dense activations in CLS-ER, the sparse coding in SCoMMER leads to the emergence of subnetworks that provide modularity and protection to other parts of the network since the entire network is not updated for each input image. This increases the robustness of the model to the class imbalance.

Overall, our method provides an effective approach to employ sparse coding in DNN and enables better utilization of long-term memory, which can effectively consolidate information across tasks and further mitigate forgetting. 

\begin{table}[tb]
\centering

\caption{\textbf{Ablation Study:} Effect of systematically removing different components of SCoMMER on the performance of the models on S-CIFAR10. All components contribute to the performance gain.}
\label{tab:abl}
\begin{tabular}{ccc|c}
\toprule
 Sparse & Long-Term & Semantic &  \multirow{2}{*}{Accuracy} \\ 
 Activations & Memory & Dropout & \\
 \midrule
\cmark & \cmark & \cmark & \textbf{69.19}\tiny±0.61 \\
\cmark & \cmark & \xmark & 67.38\tiny±1.51 \\
\xmark & \cmark & \xmark & 61.88\tiny±2.43 \\
\cmark & \xmark & \xmark & 49.44\tiny±5.43 \\
\xmark & \xmark & \xmark & 44.79\tiny±1.86 \\
\bottomrule
\end{tabular}
\end{table}

\section{Ablation Study}
To gain further insight into the contribution of each component of our method, we systematically remove them and evaluate the performance of the model in Table \ref{tab:abl}. The results show that all components of SCoMMER contribute to the performance gains. The drop in performance from removing semantic dropout suggests that it is effective in enforcing sparse coding on the representations of the model, which reduces the interference between tasks and allows semantically similar classes to share information. We also observe the benefits of multiple memory systems in CL. Additional long-term memory provides considerable performance improvement and suggests that the EMA of the learned synaptic weights can effectively consolidate knowledge across tasks. Furthermore, we observe that sparsity is a critical component for enabling CL in DNNs. Sparse activations alone significantly improve ER performance and also enable efficient utilization of semantic memory. We highlight that these individual components complement each other and that the combined effect leads to the observed performance improvement in our method. 

\section{Characteristics Analysis}
We look at different characteristics of the model to understand what enables the performance gains in our method. We analyze the models trained on S-CIFAR100 with a buffer size of 200.

\begin{figure*}[t]
        \centering
        \includegraphics[width=.95\linewidth]{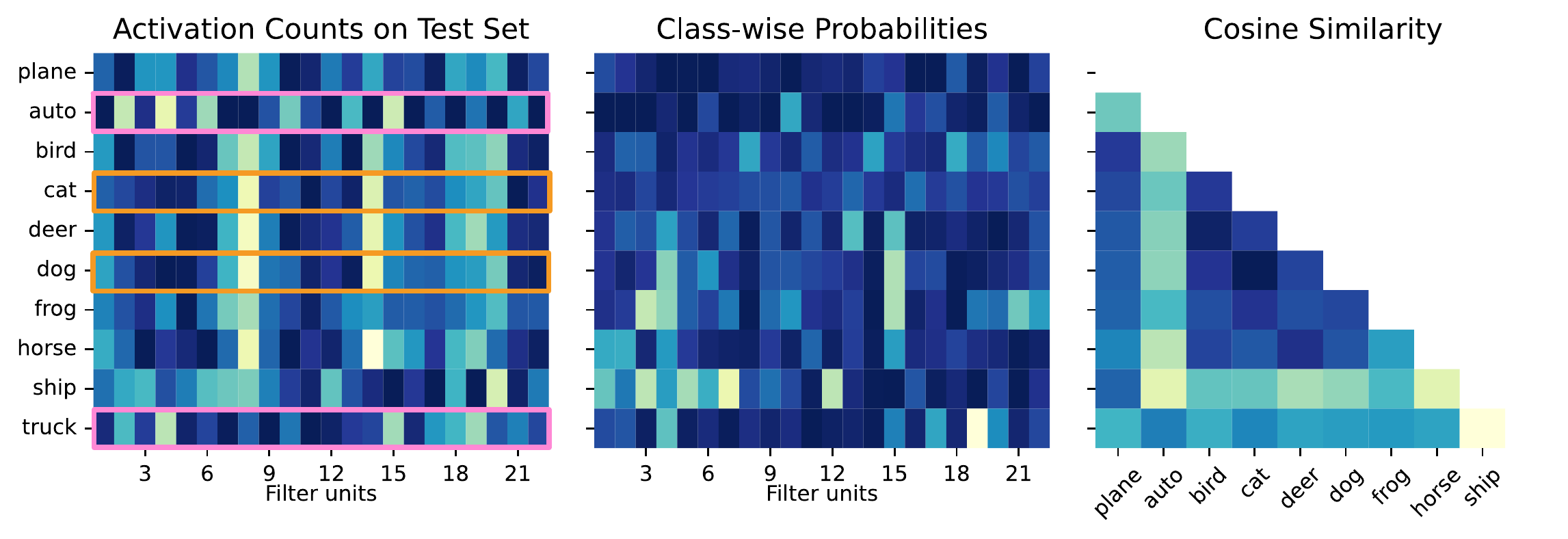}
        \caption{Class-wise activation counts of the filters in the penultimate layer of the model trained on S-CIFAR10 with 200 buffer size. Comparison of the activation counts on the test set with the learned class-wise probabilities, $P_s$, during training shows the effectiveness of semantic dropout in enforcing sparse coding. Right plot shows the cosine similarities between the activation counts of different classes. Semantically similar classes have higher correlation in activations. Darker color shows higher values.}
        \label{fig:activations}
\end{figure*}

\subsection{Stability-Plasticity Dilemma}
To better understand how well different methods maintain a balance between stability and plasticity, we look at how task-wise performance evolves as the model learns tasks sequentially. The diagonal of the heatmap shows the plasticity of the model as it learns the new task, whereas the difference between the accuracy of the task when it was first learned and at the end of the training indicates the stability of the model. Figure~\ref{fig:task_perf} shows that SCoMMER is able to maintain a better balance and provides a more uniform performance on tasks compared to baselines. While CLS-ER provides better stability than DER++, it comes at the cost of the model's performance on the last task, which could be due to the lower update rate of the stable model. SCoMMER, on the other hand, retains performance on the earlier tasks (T1 and T2) and provides good performance on the recent task. We also compare the long-term semantic and working memory performance in Figure~\ref{fig:comp_perf}. Long-term memory effectively aggregates the learned knowledge into the synaptic weights of working memory and generalizes well across tasks. 

\subsection{Emergence of Subnetworks}
To evaluate the effectiveness of activation sparsity and semantic dropout in enforcing sparse coding in the model, we look at the average activity of the units in the penultimate layer. The emerging sparse code for each class is tracked during training using the class-wise activity counter and enforced using semantic dropout probabilities (Equation~\ref{eq:semantic_dropout}). Given a test sample from class c, ideally, we would want the model to use the subset of neurons that had higher activity for the training samples from class c without providing any task information. Concretely, we track the class-wise activity on the test set and plot the normalized activation counts for a set of neurons next to their class-wise probabilities at the end of training. Figure \ref{fig:activations} shows a high correlation between the test set activation counts and the semantic dropout probabilities at the end of training, particularly for recent classes. The activation counts also hint at the natural emergence of semantically conditioned subnets, as the model utilizes a different set of units for different classes.
Furthermore, we observe that semantically similar classes have a higher degree of correlation between their activation patterns. For instance, cat and dog share the most active neurons, a similar pattern is observed between horse and deer, and car and truck. The cosine similarities between the activation counts of the different classes further supports the observation. This is even more remarkable given that these classes are observed in different tasks, particularly for cars and trucks, which are observed in the first and last tasks.

\begin{figure}[t]
        \centering
        \includegraphics[width=.95\linewidth]{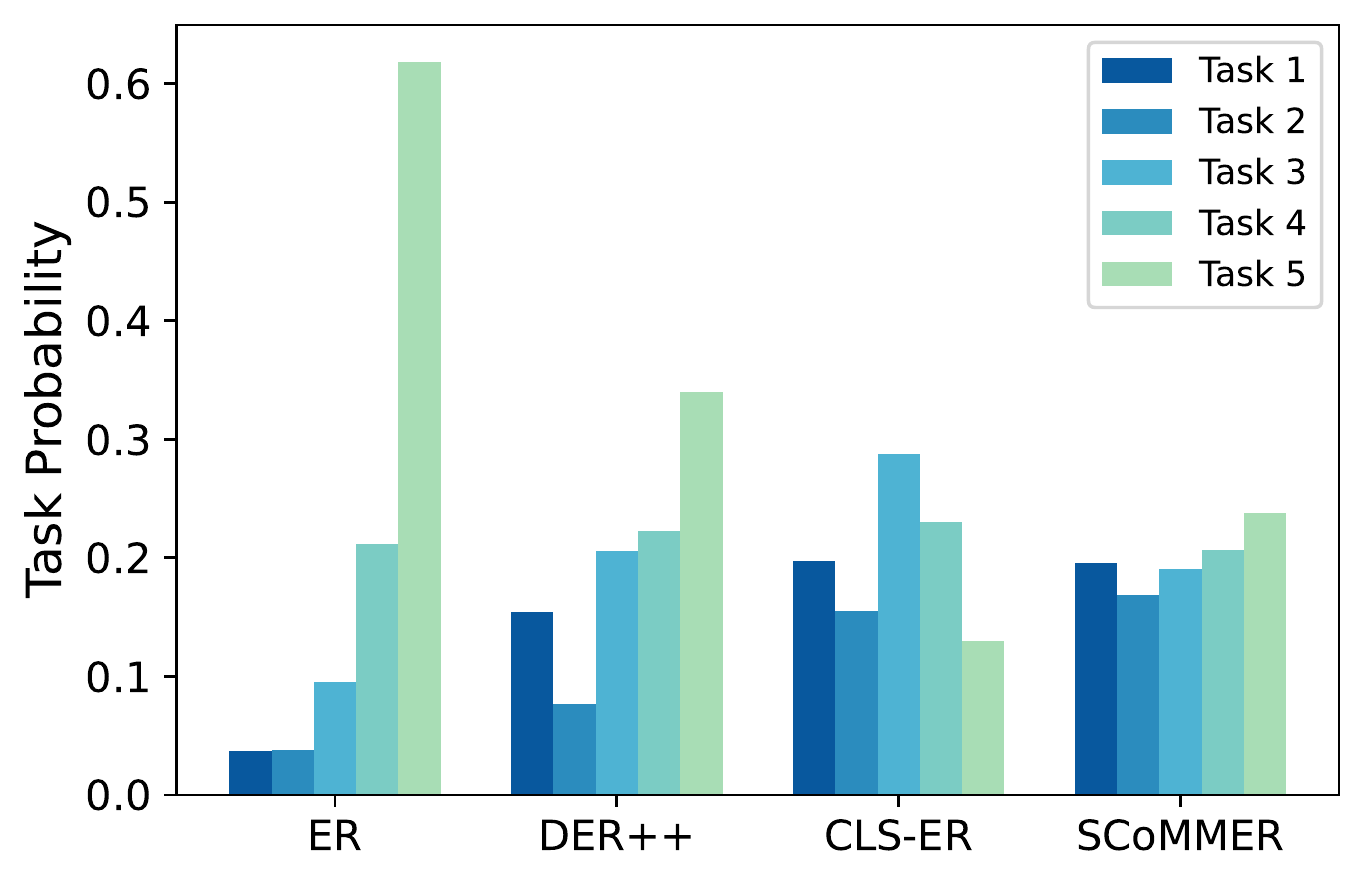}
        \caption{Average probabilities of predicting classes from each tasks at the end of training. SCoMMER provides more uniform probabilities across the tasks.}
        \label{fig:task_prob}
\end{figure}

\subsection{Task Recency Bias} 
A major challenge in CL is the recency bias, in which the update of the model on new task samples biases its predictions toward the current task~\citep{wu2019large}. This leads to considerable forgetting of earlier tasks. To compare the degree to which SCoMMER tackles this issue, we evaluate the probabilities of predicting each task by aggregating the softmax output of samples from the test set of all seen tasks and averaging the probabilities of classes in each task. Figure \ref{fig:task_prob} shows that SCoMMER provides more uniform probabilities to predict each task. CLS-ER is able to mitigate the bias towards the last task, which can be attributed to the aggregation of knowledge in the semantic memories; however, CLS-ER reduces the probability of predicting the last task, which explains the low performance. SCoMMER effectively mitigates recency bias and provides uniform prediction probabilities across tasks without any explicit regularization.

\section{Conclusion}
Motivated by the mechanisms for information representation and utilization of multiple memory systems in the brain, we proposed a novel approach to employ sparse coding in multiple memory systems. SCoMMER enforces activation sparsity along with a complementary semantic dropout mechanism, which encourages the model to activate similar units for semantically similar inputs and reduce the overlap with dissimilar inputs. Additionally, it maintains long-term memory, which consolidates the learned knowledge in working memory. Our empirical evaluation shows the effectiveness of the approach in mitigating forgetting in challenging CL scenarios. Furthermore, sparse coding enables efficient consolidation of knowledge in the long-term memory, reduces the bias towards recent tasks, and leads to the emergence of semantically conditioned subnetworks. We hope that our study inspires further research in this promising direction.


\bibliography{egbib.bib}



\newpage

\appendix

\setcounter{figure}{0}
\makeatletter 
\renewcommand{\thefigure}{S\@arabic\c@figure}
\makeatother
\setcounter{table}{0}
\makeatletter 
\renewcommand{\thetable}{S\@arabic\c@table}
\makeatother

\clearpage

\twocolumn[\section{\huge{Appendix}}]

\hspace{2mm}\newline

\section{Experimental Setting}
For a fair comparison with different CL methods in uniform experimental settings, we extended the Mammoth framework \citep{buzzega2020dark}. To disentangle the performance improvement of the algorithm from the training regimes~\cite{mirzadeh2020understanding}, we use the same network (ResNet-18), optimizer (SGD), batch size for task data and memory buffer (32), data augmentations (random crop and random horizontal flip), and the number of epochs (50) for all our experiments. 

For hyperparameter tuning, we use a small held-out validation set and perform a grip search on activation sparsity, $\gamma$, dropout strengths, $\pi_h$ and $\pi_s$, and the update frequency for long-term memory $r$. Table \ref{tab:hyperparams} provides the selected hyperparameters for each setting. Note that our method does not require an extensive hyperparameter search for different buffer sizes, and sensitivity to hyperparameters section shows that the different parameters are complementary in nature and the model performs well for a number of different combinations. Therefore, majority of parameters can be fixed, which reduces the search space of hyperparameters significantly. We report the average and one standard deviation of three different seeds.

\begin{table*}[ht]
\caption{Selected parameters for SCoMMER. For each of our experiments, we apply Heterogeneous and Semantic dropout only on the output of the last residual block in ResNet-18, the decay parameter for long-term memory is set to 0.999, the batch size of 32 is used for both the current task and the memory buffer, and the models are train for 50 epochs. For the first three ResNet blocks, we use an activation sparsity of 0.9 and vary the sparsity ratio for the last block.}
\label{tab:hyperparams}
\centering
\begin{tabular}{lc|cccccc}
\toprule
Dataset & \begin{tabular}[c]{@{}c@{}}Buffer\\ size\end{tabular} & \begin{tabular}[c]{@{}c@{}} Activation\\ Sparsity \end{tabular} & $\eta$ & $\pi_{h}$ & $\pi_{s}$ & $\gamma$ & $r$ \\ \midrule

\multirow{2}{*}{S-CIFAR10} 
 & 200 & 0.8 & 0.1 & 0.5 & 2.0 & 0.15 & 0.5 \\
 & 500 & 0.8 & 0.1 & 0.5 & 2.0 & 0.15 & 0.7 \\ \midrule

\multirow{2}{*}{S-CIFAR100} 
 & 200 & 0.9 & 0.1 & 0.5 & 3.0 & 0.15 & 0.1 \\
 & 500 & 0.9 & 0.1 & 0.5 & 3.0 & 0.15 & 0.1 \\ \midrule

\multirow{2}{*}{GCIL - Unif} 
 & 200 & 0.9 & 0.05 & 0.5 & 3.0 & 0.2 & 0.6 \\
 & 500 & 0.9 & 0.05 & 0.5 & 3.0 & 0.2 & 0.6 \\ \midrule

\multirow{2}{*}{GCIL - Longtail} 
 & 200 & 0.9 & 0.05 & 0.5 & 2.0 & 0.2 & 0.5 \\
 & 500 & 0.9 & 0.05 & 0.5 & 3.0 & 0.2 & 0.6 \\ \bottomrule
\end{tabular}
\end{table*}

\section{Continual Learning Datasets}
\label{cl-settings}
We consider the Class-IL and Generalized Class-IL setting for our empirical evaluation to extensively assess the versatility of our approach. Here, we provide details of the datasets used in each of the settings. 

\subsection{Class Incremental Learning (Class-IL)}
Class-IL~\cite{van2019three} requires the agent to learn a new disjoint set of classes with each task, and the agent has to distinguish between all the classes seen so far without the availability of task labels at the test time. We consider the split variants of the benchmark datasets S-CIFAR10 and S-CIFAR100 where the classes are split into 5 tasks with 2 and 20 classes each, respectively. The order of the classes in the experiments remains fixed, whereby for CIFAR10 the first task includes the first two classes, and so forth. 

\subsection{Generalized Class Incremental Learning (GCIL)}
GCIL~\cite{mi2020generalized} extends the Class-IL setting to more realistic scenarios. In addition to avoiding forgetting, the model has to tackle the challenges of class imbalance, learning an object over multiple recurrences. GCIL utilizes probabilistic modeling to sample three characteristics of a task: the number of classes, the classes that appear, and their sample sizes. Similarly to~\cite{arani2021learning}, we consider GCIL on the CIFAR100 dataset with 20 tasks, each with 1000 samples, and the maximum number of classes in a single task set to 50. To disentangle the effect of class imbalance from

\hspace{2mm}\newline

\noindent the ability of the model to learn from recurring classes under non-uniform task lengths, we evaluate the model on uniform (Unif) and longtail data distributions. we set the GCIL dataset seed to 1993 for all the experiments.

\section{Implementation Details}
Here, we provide more details on the implementation of k-WTA activation for CNNs and the proposed semantic dropout mechanism. 

\section{k-WTA for Convolutional Neural Networks}
The common implementation of k-WTA in convolutional neural networks involves flattening the activation map into a long $CHW \times 1$ vector and applying the activation of k-WTA in a way similar to that of the fully connected network~\cite{xiao2019enhancing,ahmad2019can}. This translates to setting some spatial dimensions of a filter to zero while propagating others. However, this implementation does not take into account the functional integrity of an individual convolution filter as an independent feature extractor and does not enable the formation of task-specific subnetworks with specialized feature extractors. Different tasks cannot utilize a different subset of filters, and we cannot track the activity of an individual filter.

Our implementation, on the other hand, assigns an activation score to each filter in the layer by taking the absolute sum of the corresponding activation map.
Given the activation map of the layer $l$, $\mathcal{A}^l$, ($C \times W \times H)$ where $C$ is the number of filters, $W$ and $H$ are the width and height, we flatten the spatial dimensions, $C \times WH)$, and the activation score for each filter $j$ is given by the absolute sum of its activations, $[C_{score}]_j$ = $\sum_{i=1}^{HW} |[\mathcal{A}^l]_{j,i}|$. We then find the value $k$ for the layer using the activation sparsity (\% of the active filters in the layer), $k \leftarrow \%k \times N_{filters}^l$ where $N_{filters}^l$ is the number of filters in the layer $l$. The $k^\text{th}$ highest value of the filter activation scores vector, $C_{score} \in \mathbb{R}^{C \times 1}$ gives the threshold value used to apply a mask to the input activation map, which only propagates the activations of filters with a score above threshold by setting the others to zero. Finally, the ReLU activation function is applied to the masked activations. Algorithm~\ref{algo:kwta} provides more details.

For the ResNet-18 network in our method, we set the activation sparsity for each ResNet block, for example \% k = [0.9, 0.9, 0.9, 0.8] enforces the activation sparsity of 0.9 in the first three ResNet blocks, that is 90\% of the filters in each convolutional layer are active for a given stimulus and 80\% in the convolutional layers of the last ResNet block.

\begin{algorithm}[t]
\caption{Global k-WTA for CNNs}
\label{algo:kwta}
\begin{algorithmic}[1]

\Statex {\bf Input:} {Activation map $\mathcal{A}$; activation ratio \%$k$; number of filters $N_{filters}$}

\Statex{}
\Statex{\textbf{Evaluate activation scores:}}
\State{Flatten the spatial dimensions:} 

\State{$C_{score} \leftarrow Reshape(C_{score}, C \times HW$)}

\State{Assign score to each filter:}

\State{$C_{score}$ = abs\_sum($C_{score}$, dim=1)}

\Statex{}
\Statex{\textbf{Calculate threshold:}}

\State{Get $k$ value for the layer:}

\State{$k \leftarrow \%k \times N_{filters}$}

\State{Return $k^\text{th}$ largest value}

\State{$C_{thresh} = k^\text{th}\_value(C_{score}, k)$}

\Statex{}
\Statex{\textbf{Mask Activation Map:}}

\State{Initialize mask with zeros}

\State{$M \leftarrow Zeros(C \times H \times W$)}

\State{Set filter mask with score above threshold to 1}

\State{$M[C_{score} > C_{thresh}] = 1$}

\State{Apply mask}

\State{$\mathcal{A} \leftarrow M \cdot \mathcal{A}$}

\Statex{}
\Statex{\textbf{Apply ReLU activation function:}}

\State{$\mathcal{A} \leftarrow \text{ReLU}(\mathcal{A})$}

\Statex \Return{$\mathcal{A}$}
\end{algorithmic}
\end{algorithm}

\subsection{Semantic Dropout}
At the beginning of training, we initialize the heterogeneous dropout probabilities $P_h$ so that for each layer $l$ the probability of ($1.1 \times \%k^l \times N_{filters}^l$) filters is set to 1 and the remaining set to 0. This is done to ensure that the learning of the first task does not utilize all filters while having the flexibility to learn a different subset of units for the classes in the first task. The semantic dropout probabilities $P_s$ are updated at the end of each epoch, once the epoch number $e$ for the task is higher than the heterogeneous dropout warm-up period $\mathcal{E}_{h}$ to allow the emergence of class-wise activity patterns before it is explicitly enforced with semantic dropout. Note that to ensure that we have enough active filters before applying k-WTA activation, when applying heterogeneous, we use the probabilities $P_h$ to sample the ($1.1 \times \%k^l \times N_{filters}^l$) filters for the given layer before applying k-WTA activation. The 1.1 factor is arbitrarily chosen and works well in practice; however, a different value can be selected. Further details of the method are provided in Algorithm~\ref{algo:sem-dropout}.

Importantly, we disable the dropout activation counter update for the buffer samples so that the sparse code is learned during task training. Also, dropout is applied only to working memory as it is learned with gradient decent, whereas the long-term memory aggregates the weights of working memory. Our analysis shows that the learned sparse coding is effectively transferred to long-term memory through ema. 

For the ResNet-18 model used in our experiments, we apply dropout at the output of each ResNet block. Although our method provides the flexibility to apply different dropout strengths for each block, we observe empirically that it works better if applied only at the output of the last ResNet block. This allows the model to learn features in the earlier layers that can generalize well across the tasks and to learn specialized features for the classes in later layers.

\begin{algorithm*}[t]
\caption{Semantic Dropout}
\label{algo:sem-dropout}
\begin{algorithmic}[1]

\Statex {\bf Input:} {Activation map $\mathcal{A}$; class labels y; activation ratio \%$k$; number of filters $N_{filters}$; dropout probabilities $P_h$ and $P_s$}

\Statex{}
\Statex{\textbf{Get the Heterogeneous Dropout Mask:}}

\State{Initialize Heterogeneous dropout mask with zeros}

\State{$H_{mask} \leftarrow Zeros(C \times H \times W$)}

\State{Calculate the sampling probabilities so that they sum to zero}

\Statex{$P_{sample}$ = $P_h$ / sum($P_h$)} 

\State{Get the indices of retained filters}

\Statex{ $N_{retain}$ = $1.1 \times \%k \times N_{filters}$}

\Statex{idx = Sample(range=$N_{filters}$, \#samples=$N_{retain}$, prob=$P_{sample}$, replace=False)} 

\State{Set the mask at retained indices to 1}

\Statex{$H_{mask}$[idx] = 1}

\Statex{}
\Statex{\textbf{Get the Heterogeneous Dropout Mask:}}

\State{Initialize Semantic dropout mask with zeros}

\State{$S_{mask} \leftarrow Zeros(C \times H \times W$)}

\State{Use the semantic dropout probabilities to select units}

\Statex{$retain = N \sim U(0, 1) \leq P_s$}

\State{Set the mask at retained indices to 1}

\Statex{$S_{mask}[retain] = 1$}

\Statex{}
\Statex{\textbf{Select the mask for each input sample}}

\State{For each sample, select Semantic dropout mask if available for the class label, otherwise use Heterogeneous dropout mask:}

\State{$M = S_{mask}$ if $P_s[y] > 0$, otherwise $H_{mask}$}

\Statex{}
\Statex{\textbf{Mask Activation Map:}}

\State{$\mathcal{A} \leftarrow M \cdot \mathcal{A}$}

\Statex \Return{$\mathcal{A}$}
\end{algorithmic}
\end{algorithm*}

\section{Performance of working memory}
To gain a better understanding of the performance of the different memories, Table~\ref{tab:wm} provides the performance of both working memory and long-term memory in the settings considered. Long-term memory consistently provides better generalization across tasks, especially in the Class-IL setting. This shows the benefits of using multiple memory systems in CL. Furthermore, it demonstrates the effectiveness of the exponential moving average of the working memory weights as an efficient approach for aggregating the learned knowledge. 

\begin{table*}[tb]
\caption{Performance of working memory and long-term memory in different settings. Long-term memory consistently provides better performance.}
\label{tab:wm}
\centering
\begin{tabular}{@{\extracolsep{4pt}}llcccccc@{}}
\toprule
\multirow{2}{*}{{Buffer}} & \multirow{2}{*}{{Memory}} & \multicolumn{2}{c}{\textbf{S-CIFAR10}} & \multicolumn{2}{c}{\textbf{S-CIFAR100}} & \multicolumn{2}{c}{\textbf{GCIL}} \\ \cmidrule{3-4} \cmidrule{5-6} \cmidrule{7-8}
 &  &  {Class-IL} & {Task-IL} &  {Class-IL} & {Task-IL} & {Unif} & {Longtail} \\ \midrule
 
\multirow{2}{*}{200} 
 & Working & 58.03\tiny±5.17 & 92.58\tiny±0.56 & 30.07\tiny±0.71 & 67.18\tiny±0.16 & 27.64\tiny±0.30 & 27.06\tiny±0.97 \\ 

 & Long-Term & \textbf{69.19}\tiny±0.61 & \textbf{93.20}\tiny±0.10 & \textbf{40.25}\tiny±0.05 & \textbf{69.39}\tiny±0.43 & \textbf{30.84}\tiny±0.80 & \textbf{29.08}\tiny±0.31 \\ 
 \midrule

\multirow{2}{*}{500} 
 & Working & 66.10\tiny±3.60 & 93.59\tiny±0.09 & 41.36\tiny±1.07 & 73.52\tiny±0.37 & 34.34\tiny±0.88 & 33.39\tiny±0.74 \\ 
 & Long-Term & \textbf{74.97}\tiny±1.05 & \textbf{94.36}\tiny±0.06 & \textbf{49.63}\tiny±1.43  & \textbf{75.49}\tiny±0.43 & \textbf{36.87}\tiny±0.36 & \textbf{35.20}\tiny±0.21 \\
\bottomrule
\end{tabular}
\end{table*}

\section{Sensitivity to Hyperparameters}
\label{cl-params}
SCoMMER employs sparse coding in a multiple-memory replay mechanism. Therefore, the setting of two sets of parameters is required: sparse coding (activation sparsity \%k and dropout strength $\pi_s$ and $\pi_h$) and the aggregation of information in long-term memory ($r$, $\alpha$). We show the effect of different sets of hyperparameters in Table~\ref{tab:hyperparam}. We can see that the different components are complementary in nature and therefore different combinations of parameters can provide similar performance. Interestingly, we observe that increasing the semantic dropout strength considerably increases the performance of the working model, but the long-term memory performance remains quite stable. The method is not highly sensitive to a particular set of parameters, and often we can fix the majority of parameters and fine-tune only a few, which significantly reduces the search space.

\begin{table}[tb]
\centering
\caption{Sensitivity to different hyperparameters. We provide the performance of Working memory and Long-term memory of models trained on S-CIFAR-10 with 200 buffer size. For all experiments $\gamma=0.15$, lr = 0.1, decay parameter = 0.999, $\pi_h=0.5$, and the model is trained for 50 epochs. For the first three ResNet blocks, we use an activation sparsity of 0.9 and vary the sparsity ratio for the last block (\%$k$)}
\label{tab:hyperparam}
\begin{tabular}{@{\extracolsep{4pt}}l|ll|cc@{}}
\toprule
\multirow{2}{*}{$r$} & \multirow{2}{*}{\%k} & \multirow{2}{*}{$\pi_s$} & \multicolumn{2}{c}{Memory} \\ \cmidrule{4-5}
& & & Working & Long-Term \\ \midrule
\multirow{9}{*}{0.4} & \multirow{3}{*}{0.7} & 1.0 & 56.65 & 69.58 \\ 
 &  & 2.0 & 59.46 & 68.30 \\
 &  & 3.0 & 59.89 & 68.93 \\ \cmidrule{2-5}
 & \multirow{3}{*}{0.8} & 1.0 & 50.25 & 67.19  \\
 &  & 2.0 & 58.01 & 69.89 \\
 &  & 3.0 & 56.91 & 68.72 \\ \cmidrule{2-5}
 & \multirow{3}{*}{0.9} & 1.0 & 51.26 & 67.49 \\
 &  & 2.0 & 56.58 & 68.32 \\
 &  & 3.0 & 56.87 & 66.89 \\ \cmidrule{1-5}
\multirow{9}{*}{0.5} & \multirow{3}{*}{0.7} & 1.0 & 57.01 & 66.80  \\
 &  & 2.0 & 59.61 & 69.26 \\
 &  & 3.0 & 60.51 & 69.00 \\ \cmidrule{2-5}
 & \multirow{3}{*}{0.8} & 1.0 & 49.09 & 67.36 \\
 &  & 2.0 & 58.03 & 69.19 \\
 &  & 3.0 & 60.37 & 67.99  \\ \cmidrule{2-5}
 & \multirow{3}{*}{0.9} & 1.0 & 49.38 & 66.27 \\
 &  & 2.0 & 60.47 & 68.16 \\
 &  & 3.0 & 57.64 & 67.88 \\ \cmidrule{1-5}
\multirow{9}{*}{0.6} & \multirow{3}{*}{0.7} & 1.0 & 56.91 & 67.85  \\
 &  & 2.0 & 61.2 & 67.64 \\
 &  & 3.0 & 62.44 & 67.94 \\ \cmidrule{2-5}
 & \multirow{3}{*}{0.8} & 1.0 & 51.11 & 65.97 \\
 &  & 2.0 & 58.61 & 66.55 \\
 &  & 3.0 & 61.01 & 69.36 \\ \cmidrule{2-5}
 & \multirow{3}{*}{0.9} & 1.0 & 49.26 & 66.93 \\
 &  & 2.0 & 58.35 & 67.44 \\
 &  & 3.0 & 60.18 & 67.90 \\
 \bottomrule
\end{tabular}
\end{table}

\end{document}